\documentclass{article}
\usepackage{ijcai24}
\usepackage{lmodern}
\usepackage{times}
\usepackage{soul}
\usepackage{url}
\usepackage[hidelinks]{hyperref}
\usepackage[utf8]{inputenc}
\usepackage[small]{caption}
\usepackage{graphicx}
\usepackage{amsmath}
\usepackage{amsthm}
\usepackage{booktabs}
\usepackage{algorithm}
\usepackage{algorithmic}
\usepackage[switch]{lineno}

\usepackage{multirow}
\usepackage{subfigure}
\usepackage{booktabs}
\usepackage{color, colortbl}
\usepackage[dvipsnames]{xcolor}
\usepackage{amssymb}
\usepackage{amsfonts}
\usepackage{epsfig}
\usepackage{graphicx}
\usepackage{epstopdf}
\usepackage{times}
\usepackage{xcolor}
\usepackage{soul}
\usepackage{float}
\usepackage{mathrsfs}
\usepackage{amssymb}
\usepackage{psfrag}
\usepackage{stfloats}
\usepackage{float}
\usepackage{comment}
\usepackage{array}
\usepackage{url}
\usepackage{caption}
\usepackage{hyperref}
\usepackage[flushleft]{threeparttable}
\definecolor{hhhh}{RGB}{51,107,158}
\definecolor{greyC}{RGB}{180,180,180}
\definecolor{greyL}{RGB}{235,235,235}
\definecolor{shadecolor}{rgb}{0.92,0.92,0.92}
\usepackage{enumitem}

\usepackage{color}

\definecolor{hhhh}{RGB}{51,107,158}
\hypersetup{
}
\definecolor{greyC}{RGB}{180,180,180}
\definecolor{greyL}{RGB}{235,235,235}
\definecolor{shadecolor}{rgb}{0.92,0.92,0.92}

\title{Cross-Problem Learning for Solving Vehicle Routing Problems}

\author{
Zhuoyi Lin$^1$\and
Yaoxin Wu$^{2}$\thanks{ Yaoxin Wu is the corresponding author.}\and
Bangjian Zhou$^{3}$\and
Zhiguang Cao$^4$\and
Wen Song$^5$\and
Yingqian Zhang$^2$ \And
Senthilnath Jayavelu$^1$
\affiliations
$^1$Institute for Infocomm Research, Agency for Science, Technology and Research (A*STAR)\\
$^2$Eindhoven University of Technology\\
$^3$Unidt\\
$^4$Singapore Management University\\
$^5$Shandong University
\emails
zhuoyi.lin@outlook.com,
wyxacc@hotmail.com,
Bangjian.zhou@unidt.com,
zhiguangcao@outlook.com,
wensong@email.sdu.edu.cn,
yqzhang@tue.nl,
J\_Senthilnath@i2r.a-star.edu.sg
}

\begin{document}

\maketitle

\begin{abstract}
Existing neural heuristics often train a deep architecture from scratch for each specific vehicle routing problem (VRP), ignoring the transferable knowledge across different VRP variants. This paper proposes the cross-problem learning to assist heuristics training for different downstream VRP variants. Particularly, we modularize neural architectures for complex VRPs into 1) the backbone Transformer for tackling the travelling salesman problem (TSP), and 2) the additional lightweight modules for processing problem-specific features in complex VRPs. Accordingly, we propose to pre-train the backbone Transformer for TSP, and then apply it in the process of fine-tuning the Transformer models for each target VRP variant. On the one hand, we fully fine-tune the trained backbone Transformer and problem-specific modules simultaneously. On the other hand, we only fine-tune small adapter networks along with the modules, keeping the backbone Transformer still. Extensive experiments on typical VRPs substantiate that 1) the full fine-tuning achieves significantly better performance than the one trained from scratch, and 2) the adapter-based fine-tuning also delivers comparable performance while being notably parameter-efficient. Furthermore, we empirically demonstrate the favorable effect of our method in terms of cross-distribution application and versatility. The code is available: \url{https://github.com/Zhuoyi-Lin/Cross_problem_learning}.
\end{abstract}

\section{Introduction}
Recently, there has been a growing trend toward applying neural methods based on deep learning to solve Combinatorial Optimization Problems (COPs), commonly known as Neural Combinatorial Optimization (NCO) \cite{bengio2021machine}. 
Among the studied COPs, the Vehicle Routing Problems (VRPs) are often favoured and chosen to verify the effectiveness of the NCO methods, especially the Traveling Salesman Problem (TSP) and Capacitated Vehicle Routing Problem (CVRP). 
On the one hand, VRPs are widely applied in real-world scenarios such as logistics, and drone delivery~\cite{wang2019vehicle,konstantakopoulos2022vehicle}. On the other hand, VRPs are known to be NP-complete problems, 
and many of them are challenging to be solved efficiently. With the advances of deep learning and its power to automatically learn neural heuristics, NCO methods have demonstrated notable promise against traditional heuristics~\cite{kool2018attention,kwon2020pomo,li2021learning,luo2023neural}. To further strengthen NCO methods, a number of recent endeavors have been paid to enhance generalization capabilities, which attempt to ameliorate the performance of the neural heuristics in solving the VRP instances with distributions or sizes unseen during training~\cite{geisler2022generalization,bi2022learning,qiu2022dimes}.

While gaining promising outcomes, current NCO methods generally learn a neural heuristic solely for each specific VRP, which brings about several issues. Firstly, such a learning paradigm would not be practical, since considerable models need to be trained from scratch when a number of VRPs are supposed to be solved, such as TSP and its variants like Orienteering Problem (OP) and Prize Collecting TSP (PCTSP). Secondly, the mainstream neural heuristics for VRPs are mostly developed based on heavy deep architectures like the Transformer model, and thus training neural heuristic even for a middle-sized problem may incur massive computational overhead and memory cost~\cite{kwon2020pomo,kim2022symnco,luo2023neural}. 
Lastly, the learned neural heuristic for a VRP tends to be ignored entirely when another VRP needs to be solved, with potentially transferable knowledge wasted.  In practice, many VRP variants share the same or similar problem structure (e.g., VRPs are always featured by customer locations denoted by node coordinates), and only differ in a few constraints in their mathematical formulations. 
Accordingly, the corresponding neural heuristics are only distinct in a few layers of their respective neural architectures. Hence a more efficient training paradigm could leverage common parts of neural networks to transfer learned parameters, 
e.g., the part of neural network trained for TSP could be re-used to train the neural heuristic for OP.

This paper mainly answers the question ``whether the trained neural heuristic for a VRP could benefit the training for other similar variants, and how could it be deployed?" To this end, we first pre-train a neural heuristic for TSP using the Transformer model, and then leverage the pre-trained model as the backbone to foster the learning of neural heuristics for other (relatively) complex VRP variants via fine-tuning. Specifically, we first modularize neural architectures for complex VRPs based on the backbone Transformer model for TSP, where additional lightweight modules beyond the backbone are applied to process problem-specific features in complex VRPs. Then, we directly apply the pre-trained backbone for TSP~\cite{kool2018attention,kwon2020pomo} in the training process of neural heuristics for the complex VRPs, by fully fine-tuning the backbone along with problem-specific modules. Finally, we propose three adapter-based fine-tuning methods to further increase the efficiency of parameter usage. Extensive experiments substantiate that the fully fine-tuned Transformers on complex VRPs significantly outperform Transformers trained from scratch on each of the VRPs. Adapter-based fine-tuning methods are inferior to full fine-tuning, but comparable to the ones trained from scratch, with significantly fewer trained parameters. Moreover, we empirically verify the efficacy of key designs in the fine-tuning process and the favorable effect of our method in the cross-distribution application. Notably, the proposed method is versatile enough to be used for different models to learn effective neural heuristics for VRPs. In summary, this paper contributes in the following four aspects:

\begin{itemize}[leftmargin=*]
    \item As an early attempt, we propose cross-problem learning for solving VRPs. The pre-trained backbone Transformer for a basic VRP is used to foster the training of Transformers for downstream complex VRPs through fine-tuning. 

\item We propose the modularization of Transformers for complex VRPs, which centers around the neural architecture of the backbone Transformer for TSP. The obtained problem-specific modules are used in the process of fine-tuning.


\item We develop fine-tuning methods for cross-problem learning, i.e., the full and adapter-based fine-tuning, by which we fine-tune the entire Transformer or only the lightweight problem-specific modules with small adapter networks.

\item Experiments show that the knowledge learned in the Transformer for TSP is well transferred to aid in training neural heuristics for other VRPs. While full fine-tuning gains better performance than Transformer trained from scratch, adapter-based fine-tuning methods attain comparable performance, with far fewer parameters trained and stored.
\end{itemize}

\section{Related Work}
In this section, we introduce the literature on neural heuristics for solving VRPs, and briefly discuss the pre-training-then-fine-tuning works in different fields.

\subsection{Neural Heuristics for VRP} 
Current neural heuristics for VRPs mostly learn policies to construct a solution in an autoregressive way.
Among diverse neural construction heuristics~\cite{bello2017neural,nazari2018reinforcement,dai2017learning,ma2024metabox}, the first breakthrough is made by the Transformer~\cite{vaswani2017attention}, which is cast as Attention Model (AM) to solve VRP variants such as TSP, OP and PCTSP~\cite{kool2018attention}. After that, numerous neural heuristics are proposed based on AM for solving VRPs~\cite{xin2021multi,kwon2020pomo,hottung2022efficient}. Among them, Kwon et al. significantly ameliorate AM with policy optimization with multiple optima (POMO)~\cite{kwon2020pomo}, and enlighten considerable follow-up works with Transformers to solve VRPs~\cite{kim2022symnco,kwon2021matrix,choo2022simulation}. As alternatives, some works endeavor to predict probabilities of edges in optimal solutions to VRPs, which are then leveraged to efficiently construct high-quality solutions with search algorithms such as greedy search, sampling, and Monte Carlo tree search~\cite{fu2021generalize,qiu2022dimes,hudson2022graph}. On the other hand, neural improvement heuristics aim to learn policies to iteratively improve an initial but complete solution. Inspired by traditional local search algorithms, a few works learn local operations such as 2-opt and swap to promote the solution ~\cite{wu2021learning,ma2021learning,wang2021bi,kim2022learning}. 
Inspired by specialized algorithms or solvers for VRPs, some literature attempts to learn repair or destroy operations in Large Neighborhood Search (LNS)~\cite{hottung2020neural,gao2020learn}. Deep learning is also applied to enhance solvers like Lin-Kernighan-Helsgaun (LKH)~\cite{xin2021neurolkh,kim2021learning} and Hybrid Genetic Search (HGS)~\cite{santana2022neural}. 



\subsection{Pre-Training, Fine-Tuning} 

The pre-training-then-fine-tuning paradigm has been widely verified effective in domains of Natural Language Processing (NLP)~\cite{dong2019unified,devlin2018bert}, Graph Neural Network (GNN)~\cite{qiu2020gcc,hu2020strategies}, Computer Vision (CV)~\cite{baobeit,chen2020simple}, etc. In this line of work, large models such as Transformers are first trained on basic tasks~\cite{dong2019unified,ribeiro2020beyond,hu2020strategies,you2020graph,dosovitskiy2020image,touvron2021training}. Then the pre-trained model is fine-tuned to solve downstream tasks, with the neural network adapted by introducing additional structures or layers (which possess far fewer parameters than the pre-trained model)~\cite{houlsby2019parameter,pfeiffer2021adapterfusion,zhao2020masking,kitaevreformer,hu2021lora}.
In this manner, the pre-training can be conducted on considerable unlabeled data to learn general patterns and features, which are re-used in downstream tasks. Also, the knowledge learned during pre-training is transferred to fine-tuning process to facilitate more efficient training. 



In this paper, the cross-problem learning method brings a similar pre-training-then-fine-tuning paradigm to the field of \emph{learning to route}. A Transformer is pre-trained for solving TSP by DRL with massive unlabeled data, which is then leveraged to fine-tune Transformers for downstream VRPs.

\section{Preliminaries}
\label{prelim}
In this section, we describe VRPs including TSP, OP, PCTSP and CVRP,
which are used in experiments to evaluate our method. Afterwards, we introduce the general DRL method to learn construction heuristics with Transformers. Finally, we analyze the paucity of existing neural heuristics.


\begin{figure*}[!t]
    \centering
    \subfigure[Transformer for TSP]{
    \includegraphics[width=0.485\linewidth, trim = 0cm 0cm 14cm 5cm, clip]{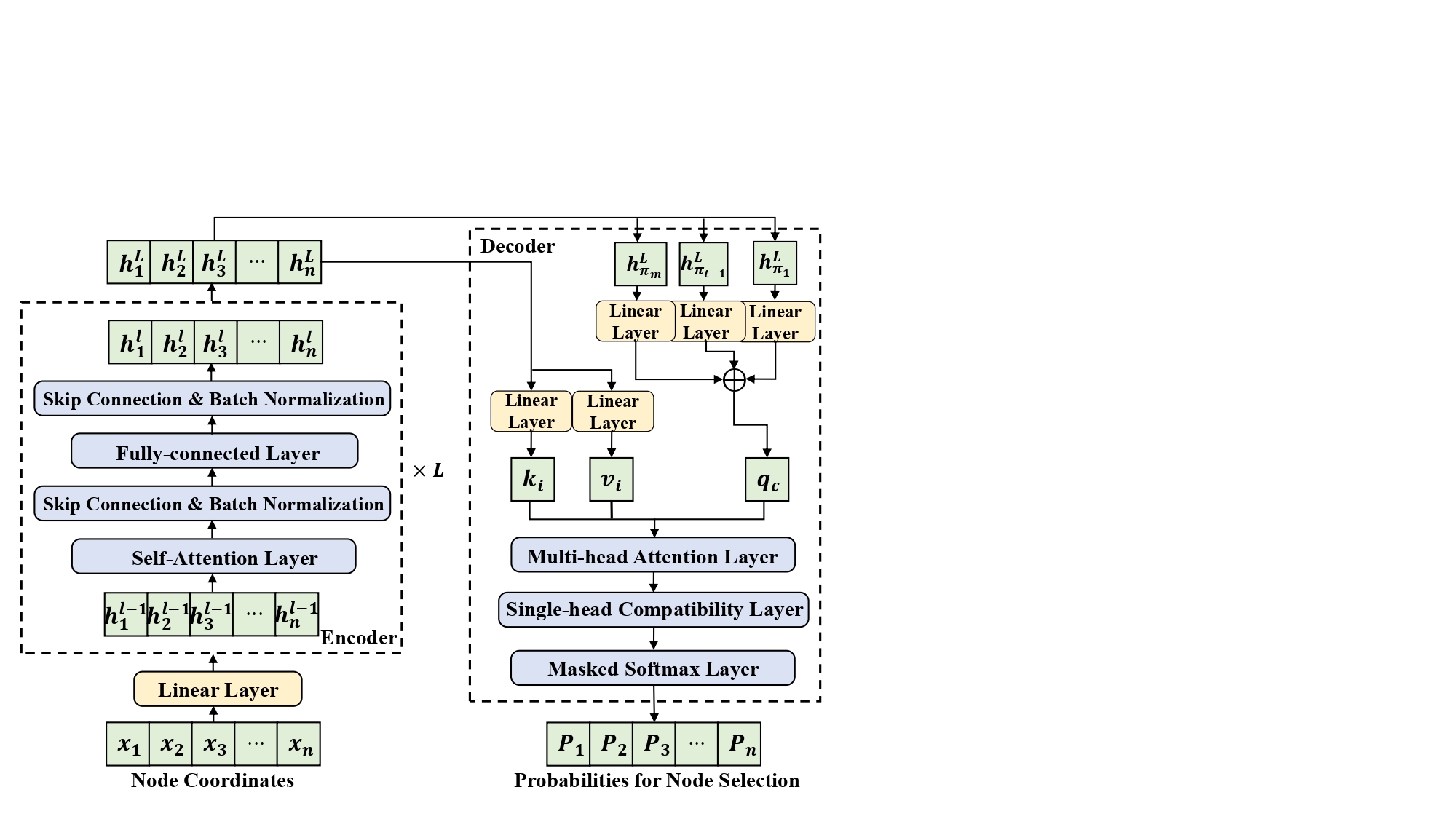}
    \label{fig1:a}
    }
    \subfigure[Transformer for OP]{
    \includegraphics[width=0.485\linewidth, trim = 0cm 0cm 14cm 5cm, clip]{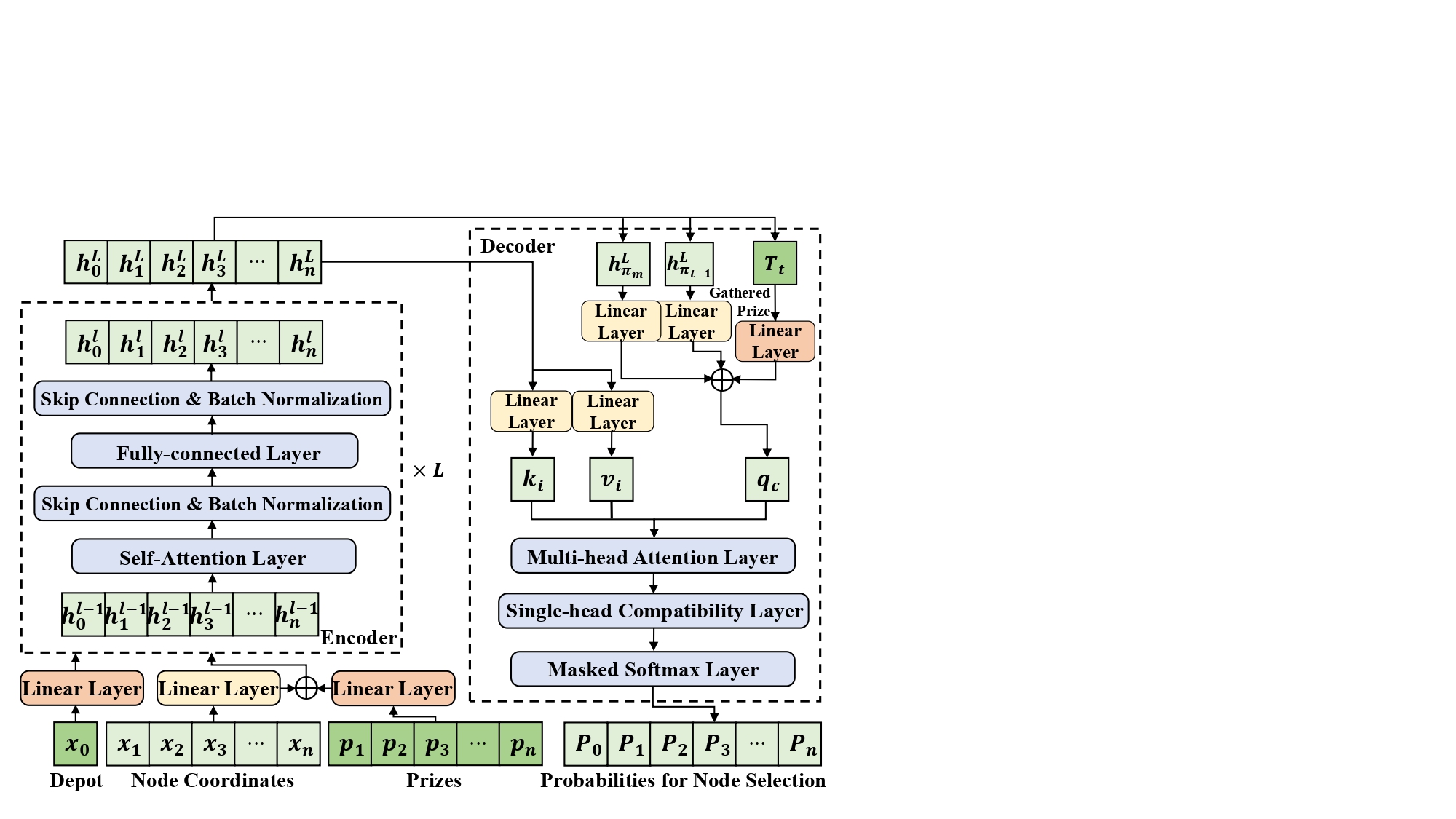}
    \label{fig1:b}
    }
    \vspace{-3mm}
    \caption{\textbf{Modularization of Transformer for OP.} Based on Transformer for TSP, we define modules for OP as three linear layers (in orange), which process problem-specific features (in dark green), i.e., depot coordinates, node prizes, total prize gathered at $t$-th step, respectively.}
    \label{fig_1}
    \vspace{-4mm}
\end{figure*}

\subsection{VRP Description} 
\label{sec:pre1}
We describe a VRP instance on a graph $\mathcal{G}=\{\mathcal{V}, \mathcal{E}\}$ with $\mathcal{V}=\{v_i\}_{i=0}^n$ signifying the node set and $\mathcal{E}=\{(v_i, v_j)|v_i, v_j\in \mathcal{V}, v_i\neq v_j\}$ signifying the edge set. Specially, $v_0$ denotes the depot if needed, e.g., in OP. On the graphs of VRP instances, nodes are often featured by coordinates $\{x_i\}_{i=0}^n$ to gain the tour length, and other problem-specific attributes (e.g. prizes $\{p_i\}_{i=1}^n$ in OP). A solution to a VRP instance is a tour $\mathbf{\pi}$, which is a sequence of nodes in $\mathcal{V}$. Given a cost/utility function $\mu(\cdot)$, the VRP instance is solved by searching the tour with the least cost or highest utility, i.e., $\pi^* = \arg\min_{\pi\in \Upsilon} \mu(\pi|\mathcal{G})\, \text{or}\, \arg\max_{\pi\in \Upsilon} \mu(\pi|\mathcal{G})$, from all feasible tours $\Upsilon$. Concretely, the cost function for TSP defines the (Euclidean) length of a tour and the utility function for OP defines the total prize of the traversed nodes in a tour. 

The solution to a VRP instance is feasible if it satisfies the problem-specific constraints. For example, a feasible solution to a TSP instance is a tour that goes through every node on the graph exactly once. A solution to OP is feasible if the length of a tour is not larger than a given value. Other VRPs like PCTSP and CVRP are defined on graphs in a similar fashion. Their details can be found in Appendix~A.\footnote{https://arxiv.org/abs/2404.11677}

\subsection{Transformer Based Construction Heuristics}
\label{sec:pre2}
Formally, the process of constructing a solution to an instance on $\mathcal{G}$ is cast as a Markov Decision Process (MDP). Given the state (i.e., the instance and partial solution) and action candidates (i.e., nodes that can be added into the partial solution), the policy (i.e., neural construction heuristic) is learned by an encoder-decoder based Transformer with parameters $\theta$. At each step of the solution construction, the Transformer takes in raw features of the state and infers probabilities of each node, by which a node is selected in a greedy or sampling way. The probabilistic chain rule of constructing a tour $\pi$ can be expressed as $p_{\theta}(\pi|\mathcal{G}) = \prod_{t=1}^{\text{T}} p_{\theta} (\pi_{t}|\mathcal{G}, \pi_{<t})$,
where $\pi_{t}$ and $\pi_{<t}$ refer to the selected node and current partial solution at step $t$. $\text{T}$ denotes the maximum number of steps. Given the reward defined as the negative cost (or utility) of a tour, the REINFORCE~\cite{williams1992simple} algorithm is often leveraged to update the policy with the following equation:
\begin{equation}
    \label{eq:reinforce}
    \nabla_{\theta} \mathcal{L}(\theta|\mathcal{G}) = \mathbb{E}_{p_{\theta}(\pi|\mathcal{G})} [(\mu(\pi)-b(\mathcal{G})) \nabla_{\theta}\log p_{\theta}(\pi|\mathcal{G})],
\end{equation}
where $\mathcal{L}(\theta|\mathcal{G})=\mathbb{E}_{p_{\theta}(\pi|\mathcal{G})}\mu(\pi)$ is the expected cost and $b(\cdot)$ is a baseline function to reduce the variance of the estimation and raise the training efficiency~\cite{sutton2018reinforcement}. 


\section{The Proposed Method}
\label{methodology}

The cross-problem learning for solving VRPs falls within the pre-training-then-fine-tuning paradigm. Given the Transformer for a basic VRP (TSP in this paper), we train it to learn neural heuristics for solving the basic VRP with a DRL algorithm. Then, the pre-trained Transformer as the backbone is applied to train neural heuristics for downstream VRPs by fine-tuning. According to the modularization of Transformers, we propose different fine-tuning methods, which fully fine-tune the backbone with problem-specific modules, or only fine-tune lightweight problem-specific modules with small adapter networks. In the below, we elaborate on the modularization of Transformers for VRPs, the pre-training method for TSP by DRL, and different fine-tuning methods in detail.

\begin{figure*}[!t]
    \centering
    \subfigure[Inside Tuning]{
       \hspace{-2.5mm}
       \includegraphics[width=0.3\linewidth, trim = 0cm 0.5cm 18cm 6.5cm, clip]{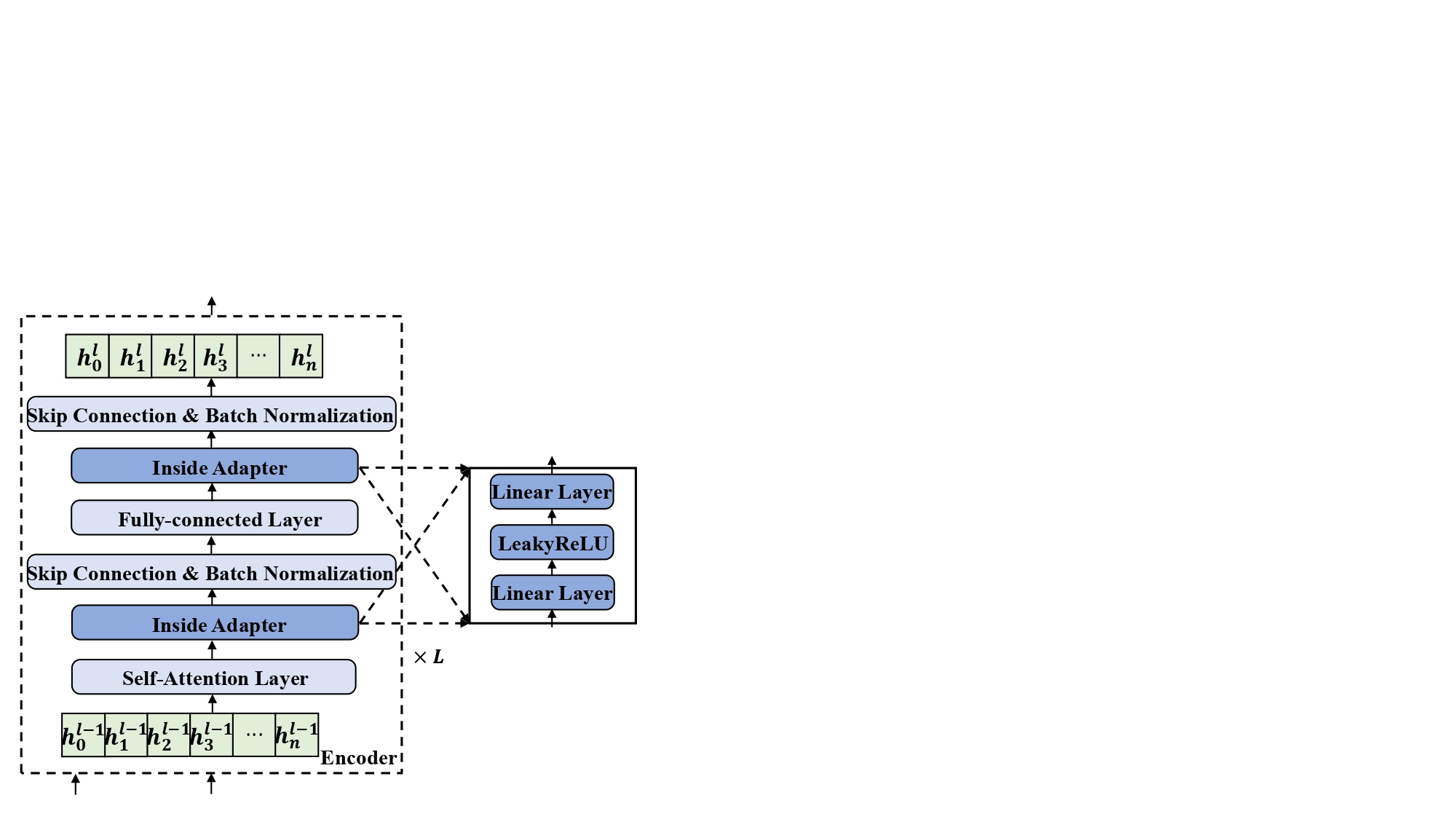}
    \label{fig0:b}
    }
    \subfigure[Side Tuning]{
   \hspace{-6mm}
\includegraphics[width=0.38\linewidth, trim = 0cm 0.1cm 13cm 6.5cm, clip]{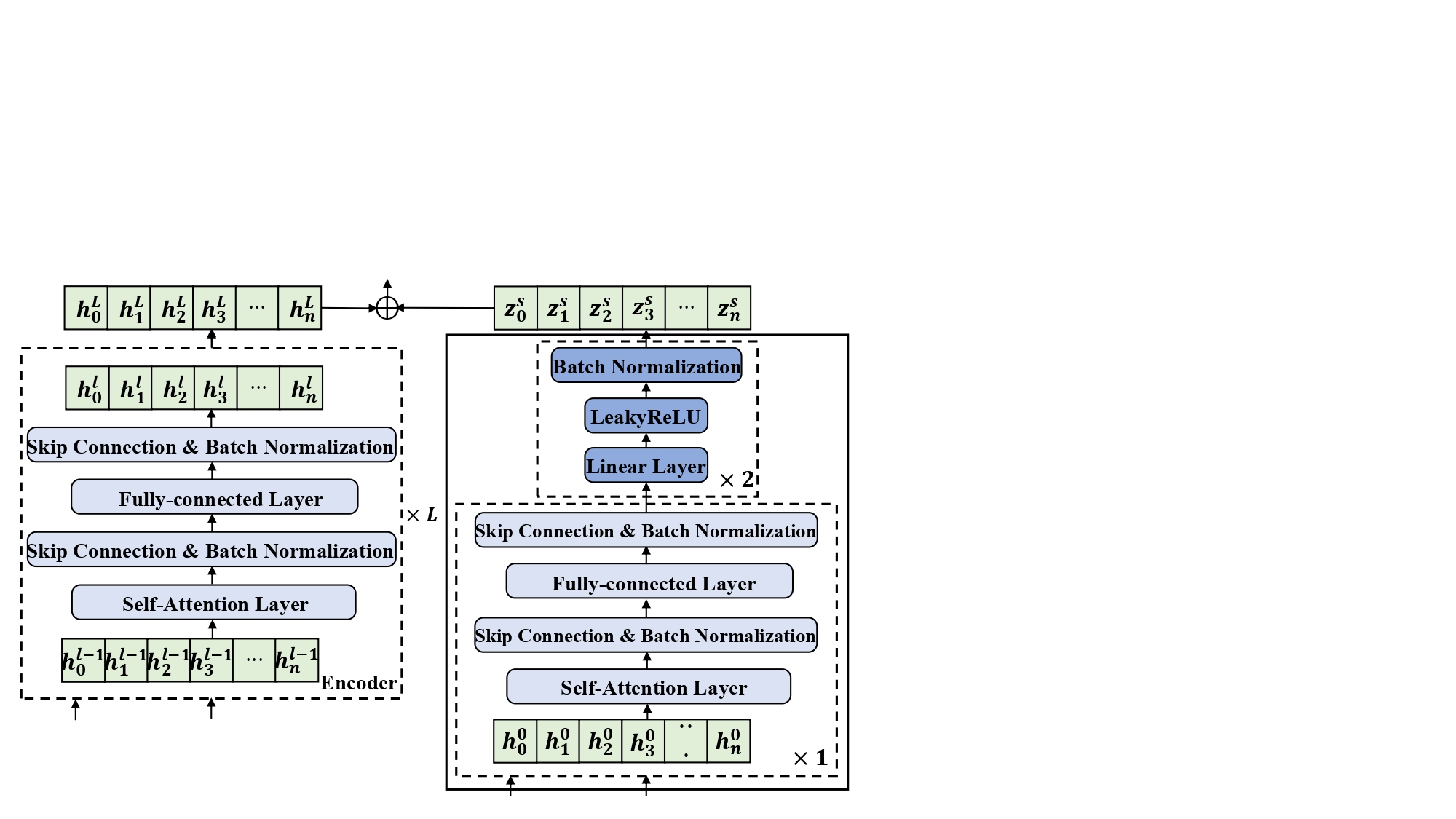}
    \label{fig0:c}
    } 
    \subfigure[LoRA]{
   \hspace{-7mm}
\includegraphics[width=0.35\linewidth, trim = 0cm 0cm 17.5cm 8cm, clip]{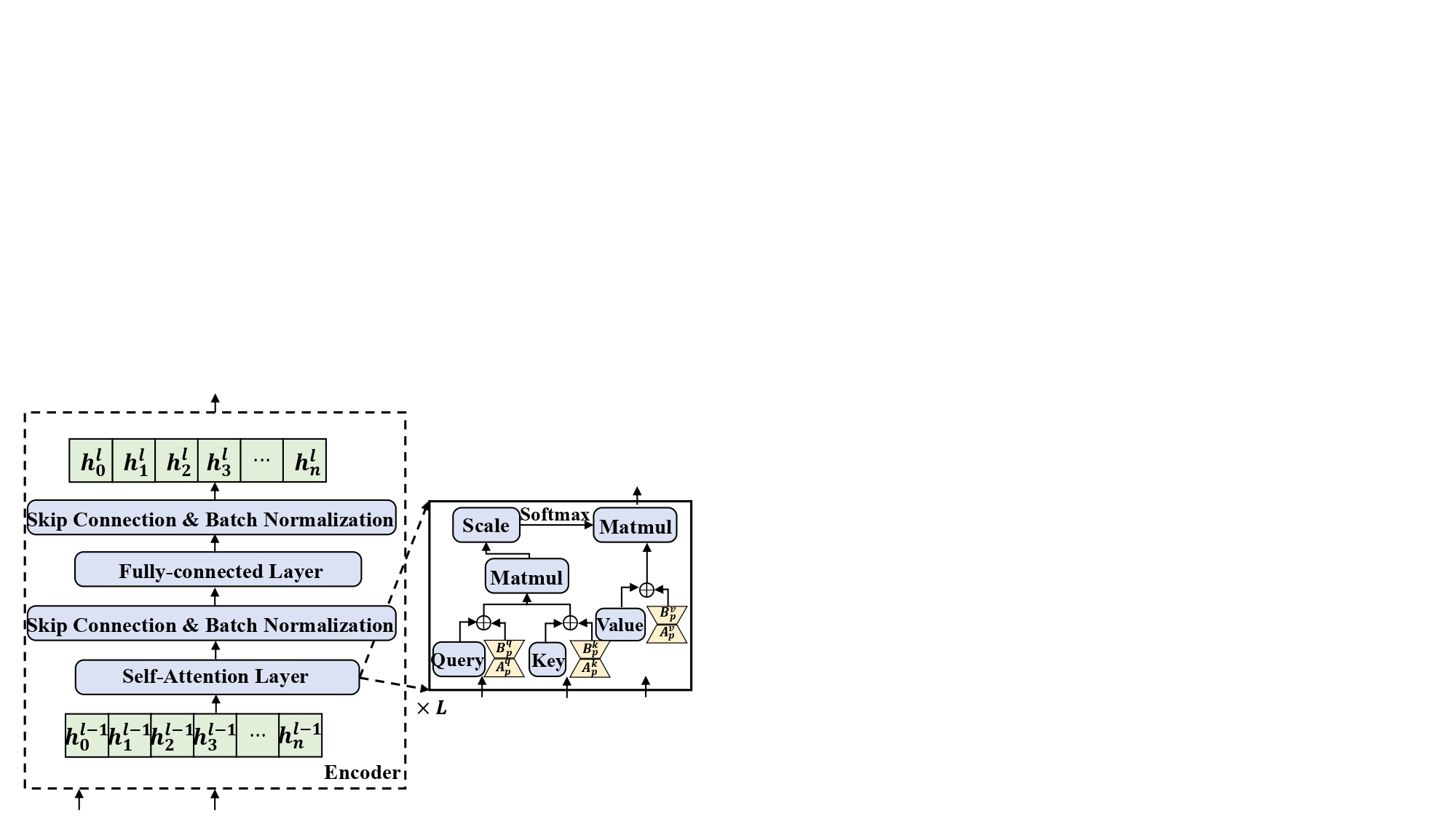}
    \label{fig0:d}
    } 
    \vspace{-4mm}    
    \caption{Three types of adapter-based fine-tuning.}
    \label{fig_0}
    \vspace{-4mm}
\end{figure*}

\subsection{Modularization of Transformers for VRPs}\label{sec:Modularization}

Most neural heuristics for VRPs are developed from Transformers~\cite{kool2018attention,kwon2020pomo,kwon2021matrix} with similar neural architectures.
To learn heuristics for TSP, the coordinates of customer nodes $\{x_i\}_{i=1}^n$ are linearly projected to initial embeddings $\{h_i^0\}_{i=1}^n$. The encoder in a Transformer processes $\{h_i^0\}_{i=1}^n$ to derive the advanced node embeddings $\{h_i^L\}_{i=1}^n$, which are then used in the decoder to output probabilities of nodes to be selected, 
as shown in Figure~\ref{fig1:a}. Based on the Transformer for TSP, we modularize neural architectures for complex VRPs into the backbone Transformer (for TSP) and problem-specific modules.

Taking OP as an example (shown in Figure~\ref{fig1:b}), the Transformer resorts to additional modules (i.e., linear layers) to process the coordinate of depot $x_0$ and prizes on customer nodes $\{p_i\}_{i=1}^n$, respectively. The linearly transformed prizes are added to the initial embeddings of coordinates to complement the information on customer nodes in OP. Then the resulting embeddings $\{h_i^0\}_{i=1}^n$ are concatenated with the initial embedding of the depot node $h_0^0$, which together are taken as input to the encoder. Similarly, the decoder uses one additional module (i.e., linear layer) to inject the dynamic feature, i.e.,  the remaining max length $T_t$ at the $t$-th step, into the query vector for the subsequent computation. Note that we also dump the unnecessary information in TSP from the query embedding, e.g., the embedding of the firstly selected node, which is always the depot and thus senseless in OP.

Consequently, the Transformer for OP is modularized into the backbone Transformer (for TSP) plus the three additional linear layers to process additional features. Similarly, we can modularize Transformers for other VRPs such as PCTSP and CVRP. Taking PCTSP as an example, we only need modules to transform 1) the depot's coordinate, prizes, and penalties before the encoder, and 2) the remaining prize in the decoder.

\subsection{Pre-Training on TSP}
Given the above modularization, we propose to first train the backbone Transformer for TSP, and then fine-tune the problem-specific modules for downstream tasks. The rationale behind the use of TSP as the basic task for pre-training is that 1) as showed in the modularization, the Transformer for TSP is included in Transformers for other VRPs (e.g., OP, PCTSP);
2) as a basic VRP, TSP is merely featured by node coordinates, which are also included in other VRPs. 
Hence, we can leverage the backbone Transformer for TSP to learn useful node representations to reflect the common knowledge in VRPs (e.g., node locations and their distances), and then insert the backbone in Transformers for downstream VRPs.

In this paper, we take Transformer based AM as the backbone to pre-train the neural heuristic for TSP, since AM is able to solve a wide range of VRPs with strong performance. 
We illustrate the neural architecture of AM in Figure~\ref{fig1:a}, and refer interested readers to the original work in~\cite{kool2018attention} for more details. We will focus more on the usage of the pre-trained AM in fine-tuning for downstream VRPs.



\subsection{Full Fine-Tuning for VRPs}
A straightforward fine-tuning method is to leverage the pre-trained AM for TSP as a warm start for training a Transformer for any downstream VRP. Specifically, we load the parameters of pre-trained AM to the corresponding part of the Transformer, and leave the parameters on the extra problem-specific modules randomly initialized. In this paper, we uniformly initialize their parameters within the range $(-1/\sqrt{d}, 1/\sqrt{d})$, where $d=128$ is the hidden dimension in AM. Taking the OP as an example, we apply three linear layers $LL_k(z) = \mathbf{W}_k\cdot z+\mathbf{b}_k$, $k=\{0,1,2\}$ to process the coordinate of depot node, prizes of customer nodes, and total prize gathered, respectively. The trainable parameters $\mathbf{W}_0\in\mathbb{R}^{2\times d}$; $\mathbf{W}_1, \mathbf{W}_2\in\mathbb{R}^{1\times d}$; $\mathbf{b}_0, \mathbf{b}_1, \mathbf{b}_2\in\mathbb{R}^{d}$ are uniformly initialized.
During training, we tune parameters in both the backbone and additional problem-specific modules with the same optimizer and learning rate, following the  REINFORCE mentioned in Section~\ref{sec:pre2}. We empirically observe that such simple fine-tuning is able to significantly increase the training efficiency, in comparison to existing methods that train neural heuristics from scratch for each VRP.

\subsection{Adapter-Based Fine-Tuning for VRPs}
Despite the handy implementation of full fine-tuning, it still depends on massive parameters to be trained for every downstream VRP. Given a broad spectrum of VRP variants, it would not be realistic to train and store many heavy models with limited memory resources. An ideal alternative is that we only store problem-specific modules along with small networks for a VRP, which can be used with the backbone (for TSP) to assemble the Transformer for solving it. 
To this end, we provide three adapter-based fine-tuning methods, i.e., inside tuning, side tuning and LoRA. For each downstream VRP, we only train its problem-specific modules and small adapter networks, while freezing the parameters of the pre-trained backbone. 
Despite similar adaptation techniques in other domains~\cite{zhang2020side,sunglst,hu2021lora}, we take an early step to propose three fine-tuning methods with adapter networks for solving VRPs.

The adapter network in the inside tuning comprises two linear layers with a LeakyReLU (LR) activation function in between such that:
\begin{equation}
    \label{eq:inside}
    \text{APT}_{\text{in}}(h_i) = \mathbf{W}_1^{\text{in}} \cdot \text{LR}(\mathbf{W}_0^{\text{in}}\cdot h_i+\mathbf{b}_0^{\text{in}})+\mathbf{b}_1^{\text{in}},
\end{equation}
where $i\in \{0,\ldots,n\}$ is the node index. We set the trainable parameters $\mathbf{W}_0^{\text{in}}\in\mathbb{R}^{d\times (d/2)}$; $\mathbf{W}_1^{\text{in}}\in\mathbb{R}^{(d/2)\times d}$; $\mathbf{b}_0^{\text{in}} \in\mathbb{R}^{d/2}$; $\mathbf{b}_1^{\text{in}} \in\mathbb{R}^{d}$, which are uniformly initialized with the range $(-1/\sqrt{d}, 1/\sqrt{d})$. We put the above adapter network right after the self-attention layer and fully-connected layer in the backbone Transformer's encoder, as shown in Figure~\ref{fig0:b}. 

Instead of adjusting intermediate embeddings in the encoder, the side tuning adjusts the output from the encoder, with the following neural architecture: 
\begin{equation} \label{eq:side1}
h^{'}_i=\mathbf{BN}\left(h_i+\mathbf{MSL}_i\left(h_0,\ldots,h_n\right)\right),
\end{equation}
\begin{equation} \label{eq:side2}
h_i^{''}=\mathbf{BN}\left(h^{'}_i+\mathbf{FL}(h^{'}_i)\right),
\end{equation}
\begin{equation} \label{eq:side3}
\text{APT}_{\text{si}}(h_i^{''}) = \mathbf{BN}(\text{LR}(\mathbf{W}_1^{\text{si}}\cdot \mathbf{BN}(\text{LR}(\mathbf{W}_0^{\text{si}}\cdot h_i^{''}+\mathbf{b}_0^{\text{si}}))+\mathbf{b}_1^{\text{si}})),
\end{equation}
where $i\in \{0,\ldots,n\}$ is the node index. In Eqs. (\ref{eq:side1}) and (\ref{eq:side2}), we use the similar neural structure in the encoder of AM (once rather than $L$ times), where it processes node embeddings by Multi-head Self-attention Layer ($\mathbf{MSL}$) and Fully-connected Layer ($\mathbf{FL}$), with either followed by skip connection and batch normalization ($\mathbf{BN}$). More details of these components can be found in~\cite{kool2018attention}. In Eq.~(\ref{eq:side3}), the node embeddings are further evolved by a linear layer, the LeakyReLU activation and batch normalization, where $\mathbf{W}_0^{\text{si}}, \mathbf{W}_1^{\text{si}}\in\mathbb{R}^{d\times d}$; $\mathbf{b}_0^{\text{si}},\mathbf{b}_1^{\text{si}} \in\mathbb{R}^d$ are trainable parameters and uniformly initialized. This adapter network is put beside the encoder and evolves the initial node embeddings $\{h_i^0\}_{i=0}^n$ into $\{z_i^s\}_{i=0}^n$, which are added into the output from the encoder (with frozen backbone weights), as shown in Figure~\ref{fig0:c}.

Inspired by Low-Rank Adaptation (LoRA) in~\cite{hu2021lora}, we design a low-rank decomposition to adjust the output from any matrix in the pre-trained Transformer,
\begin{equation}  \label{eq:lora}
\text{APT}_{\text{lo}}(h_i)=\mathbf{W}_p\cdot h_i+\mathbf{B}_p\mathbf{A}_p\cdot h_i,
\end{equation}
where $i\in \{0,\ldots,n\}$ is the node index. $\mathbf{W}_p\in\mathbb{R}^{d\times d}$ represents the pre-trained weight matrix in the backbone Transformer for TSP; $\mathbf{B}_p\in\mathbb{R}^{d\times r}$ and $\mathbf{A}_p\in\mathbb{R}^{r\times d}$ are two trainable matrices with $r=2\ll d$, which we initialize by $0$ and Gaussian(0,1). 
We apply Eq. (\ref{eq:lora}) to adjust the output from query, key, and value matrices in the encoder of pre-trained Transformer, with one example shown in Figure~\ref{fig0:d}. Meanwhile, we exert LoRA on the weight matrix in the initial linear projection for better results we can find. 


\section{Experiments}
\label{exps}
We empirically evaluate cross-problem learning from different perspectives. Firstly, we compare it with other baselines to show its performance on OP and PCTSP. Then, we ablate on the key designs in the fine-tuning methods. Finally, we demonstrate the effectiveness of our method, 
in terms of cross-distribution application and versatility.

Following the existing works, we generate the coordinates of nodes by uniformly sampling from the unit square $[0,1]\times[0,1]$. For OP, we set prizes all equal to 1. For PCTSP, we randomly sample prizes and penalties from Uniform distributions, following~\cite{kool2018attention}. 
Meanwhile, we generate different-sized instances for OP and PCTSP, by setting the number of customer nodes to $n=$ 20, 50, and 100.
We then use the Transformer based AM in~\cite{kool2018attention} to implement the cross-problem learning, and maintain most of the original experimental settings for a fair comparison. Specifically, we train the model with 2500 batches in each epoch and use 512 instances in each batch (except for the problem with $n=$ 100, where we use 256 instances). For each epoch, the training instances are generated on the fly, and we use 10,000 instances for the validation. Then the paired t-test ($\alpha=5\%$) is conducted after each epoch to replace the baseline model, according to the improvement significance. 
For all problems and sizes, we train 100 epochs for the full convergence of the training curves.
The learning rate is set as $10^{-4}$ in all our methods, as the one for the original AM\footnote{We tried to assign different learning rates to adapter networks and problem-specific modules for fine-tuning, which did not result in better performance.}. 
The server with Intel Core i9-10940X CPU and NVIDIA GeForce RTX 2080Ti is used in our experiments.


We compare our method with AM~\cite{kool2018attention}, the DRL based method with a strong performance on a variety of VRPs. We deploy the cross-problem learning based on its Transformer models. 
We also compare various baselines as follows. Regarding OP, we include 1) Gurobi, the state-of-the-art exact solver for various COPs~\cite{gurobi}; 2) Compass, the specialized Genetic Algorithm (GA) for OP~\cite{kobeaga2018efficient}; 3) Tsili, a classic randomized construction heuristic for OP with the manually crafted node probabilities ~\cite{tsiligirides1984heuristic}. As for PCTSP, besides Gurobi, we run two additional baselines: 4) ILS, the iterated local search metaheuristic that is widely used for routing problems; and 5) OR-Tools, a mature metaheuristics based solver for routing problems.


\begin{table*}[ht] 
  \begin{center}
  \begin{scriptsize}
  \renewcommand\arraystretch{0.65}
  \resizebox{\textwidth}{!}{ 
  \begin{tabular}{ll|ccc|ccc|ccc
  |c}
    \toprule
    \midrule
    \multicolumn{2}{c|}{\multirow{3}{*}{}} &
    \multicolumn{3}{c|}{$n=20$} & \multicolumn{3}{c|}{$n=50$} & \multicolumn{3}{c|}{$n=100$} & \multirow{2}*{\#Par.}\\
     & & Obj. & Gap & Time & Obj. & Gap & Time & Obj. & Gap & Time  \\
    \midrule
    \multirow{8}*{\rotatebox{90}{TSP $\rightarrow$ OP}} & Gurobi  & 10.57 & 0.00\% & 4m &  & -- &  &  & -- & & -- \\
    & Gurobi (30s) & 10.57 & 0.01\% & 4m & 29.30 & 0.92\%& 1h & 43.33 & 27.00\% & 2h& -- \\     
    & Compass & 10.56 & 0.12\% & 16s & 29.58 & 0.00\% & 58s & 59.35 & 0.00\% & 3m & -- \\     
    & Tsili (Greedy)  & 8.82 & 16.58\% & 1s & 23.89 & 19.22\% & 1s & 47.65 & 19.71\% & 1s & -- \\
     & Tsili (Sampling)$^{*}$  & 10.48 & 0.85\% & 28s & 28.26 & 4.46\% & 2m & 54.27 & 8.56\% & 6m & -- \\
         \cmidrule(lr){2-12}
     & AM (Greedy) &10.28 & 2.68\% & 9s & 28.28 & 4.39\% & 9s & 55.94 & 5.75\% & 11s & 694K \\
     & AM (Sampling) & 10.49 & 0.71\% & 3m & 29.30 & 0.95\% & 9m & 58.39 & 1.62\% & 19m & 694K \\
     & Ours (Full) & 10.49 & 0.76\% & 3m & \textbf{29.46} & \textbf{0.41\%} & 8m & \textbf{58.88} & \textbf{0.79\%} & 19m & 694K \\
     & Ours (Adapter) & \textbf{10.50} & \textbf{0.69\%} & 3m & 29.43 & 0.51\% & 8m & 58.76 & 0.99\% & 19m & \textbf{198K} \\
    \midrule\midrule
    
    \multirow{8}*{\rotatebox{90}{TSP $\rightarrow$ PCTSP}}  
    & Gurobi & 3.13 & 0.00\% & 1m && --&  & & -- & & -- \\       
    & Gurobi (30s) & 3.13 & 0.00\% & 1m & 4.48 & 0.03\% & 52m &  & -- & & -- \\
     & ILS (C++)$^{*}$  & 3.16 & 0.77\% & 16m & 4.50 & 0.36\% & 2h & 5.98 & 0.00\% & 12h & --\\
     & OR-Tools (60s) & 3.13 & 0.01\% & 5h & 4.48 & 0.00\% & 5h & 6.07 & 1.56\% & 5h & -- \\
              \cmidrule(lr){2-12}
     & AM (Greedy) & 3.19 & 1.81\% & 10s & 4.60 & 2.68\% & 10s & 6.25 & 4.52\% & 8s & 694K \\
     & AM (Sampling) & 3.15 & 0.72\% & 3m & 4.52 & 0.88\% & 8m & 6.08 & 1.67\% & 19m & 694K \\
     & Ours (Full) & \textbf{3.15} & \textbf{0.58\%} & 3m & \textbf{4.51} & \textbf{0.60\%} & 8m & \textbf{6.04} & \textbf{1.00}\% & 19m & 694K \\
     & Ours (Adapter) & 3.15 & 0.64\% & 3m & 4.51 & 0.71\% & 8m & 6.06 & 1.34\% & 20m & \textbf{199K} \\
    \midrule
    \bottomrule
  \end{tabular}}
	\begin{tablenotes} \small
		\item[\textbf{1}] \textbf{Bold} means the best results among learning based methods. $*$ means the results of the method reported in~\cite{kool2018attention}.
	\end{tablenotes}
 \end{scriptsize}
  \end{center}
    \vskip -0.15in
    \caption{Comparison between cross-problem learning and baselines.}
  \vskip -0.05in
    \label{exp_1}
\end{table*}


\begin{figure*}[!ht]
\centering
\begin{tabular}{@{}c@{}c@{}c@{}c@{}}
  {\includegraphics[width=0.33\linewidth]{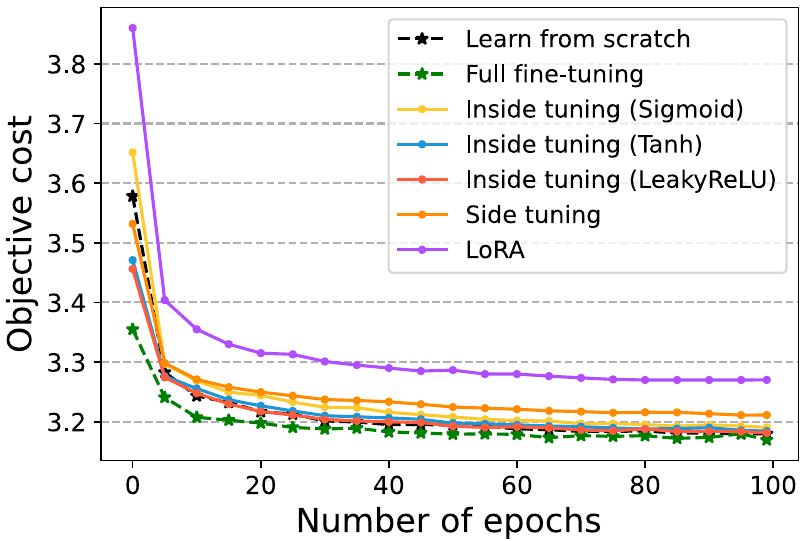}}&
  {\includegraphics[width=0.33\linewidth]{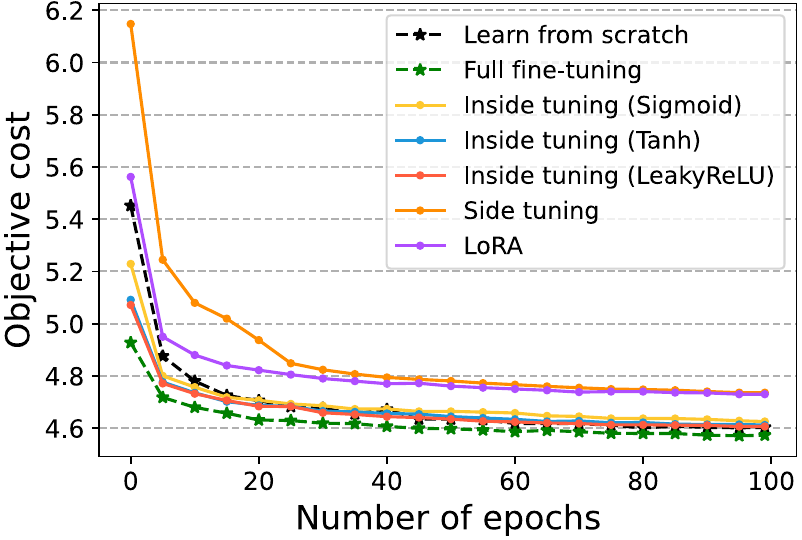}}&
  {\includegraphics[width=0.33\linewidth]{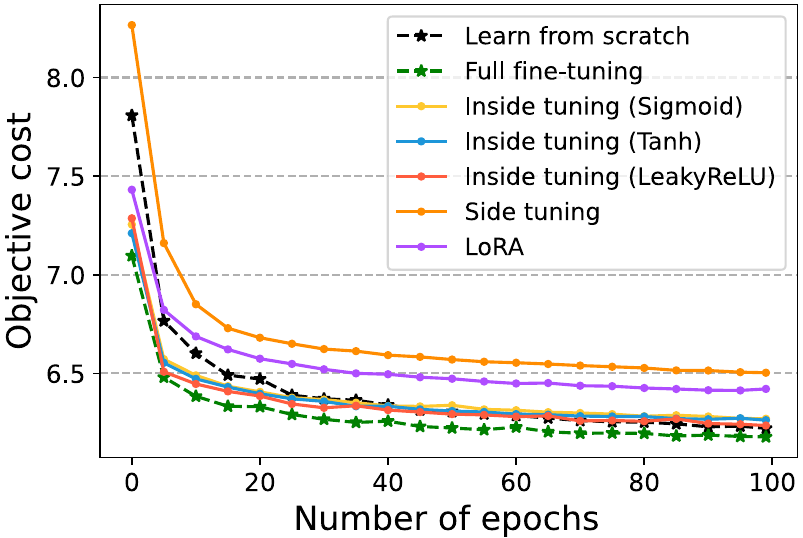}}&
\end{tabular}
\vspace{-3mm} 
\caption{Ablation study on fine-tuning methods and activation functions on PCTSP ($n=20, 50, 100$), respectively.}
\label{fig:ablation} 
\vspace{-4mm}
\end{figure*}

\subsection{Comparison}
We test all methods with 10,000 instances of OP and PCTSP, two typical VRPs that aim to maximize the total prize and minimize the distance plus the total penalty in the tour, respectively. We report the results of AM and Tsili by two decoding strategies, i.e., greedy search and sampling. For our method, we only report the better results derived by sampling. The sampling size is 1,280 for these methods, following~\cite{kool2018attention}. For adapter-based fine-tuning, we only report results derived by inside tuning, since it significantly outstrips side tuning and LoRA (see Section~\ref{sec:abl}). We compare the methods by different metrics, including the average objective value, average gap (to the best solutions), total runtime, and the number of trained parameters (for learning based methods). In Table~\ref{exp_1}, we can observe that our method with either full fine-tuning or inside tuning consistently outperforms AM with either greedy search or sampling, with significantly smaller gaps and similar runtime. While Gurobi achieves the optimal solutions for small problems, it cannot derive solutions for middle-sized OP and PCTSP ($n$=50, 100). Instead, we run Gurobi as a heuristic with a 30s limit, but it is inferior to our methods on OP ($n$=50, 100) or fails to obtain any solutions on PCTSP ($n$=100). While Compass and ILS are specialized algorithms for OP and PCTSP respectively, our method achieves comparable results in all cases, with most gap differences smaller than $1\%$. Moreover, our methods significantly surpass Tsili on OP ($n$=20, 50, 100), and outperform OR-Tools on PCTSP ($n$=100). Lastly, the inside tuning costs far fewer parameters for training, in contrast to AM and the full fine-tuning, which reduces considerable memory resources for solving each VRP (except the backbone Transformer for TSP). Besides transferring knowledge between problems with the same (Uniform) distribution, the cross-problem learning can effectively solve downstream VRPs with different distributions. Results on such cross-distribution applications are provided in Appendix B.


\subsection{Ablation Study}
\label{sec:abl}
In this part, we conduct an ablation study to demonstrate the impact of key designs for the proposed fine-tuning methods. Given the same pre-trained backbone, we first compare the inside tuning, the side tuning, and LoRA. Then, we show the sensitivity of the inside tuning approach by varying activation functions, including Sigmoid, Tanh, and LeakyReLU. With the changed components, we draw training curves on PCTSP ($n$=20, 50, 100) and display them in Figure~\ref{fig:ablation}. To investigate the effects of different training methods, we also add the curves for the training from scratch and the full-tuning. 
As shown, although side tuning is equipped with a larger adapter network (i.e., 332K vs. 199K), it is significantly inferior to inside tuning on all sizes. On the other hand, despite the disconcerting performance of LoRA on PCTSP ($n$=20), it achieves similar results with side tuning for $n$=50 and is significantly superior to side tuning for $n$=100 with only 127K parameters. 
In addition, the activation function influences the performance, and the inside tuning with LeakyReLU outperforms the one with Sigmoid or Tanh. Notably, the curves of the inside tuning drop faster in the early epochs than the original AM trained from scratch on PCTSP ($n$=50, 100), and the inside tuning with LeakyReLU achieves similar convergences to AM on all sizes, with far fewer trained parameters.




\begin{table*}[!t] 
\scriptsize
  \begin{center}
  \renewcommand\arraystretch{0.85}
  \resizebox{\textwidth}{!}{ 
  \begin{tabular}{l|ccc|ccc|ccc|c}
    \toprule
    \midrule
    \multicolumn{1}{c|}{\multirow{3}{*}{}} &
    \multicolumn{3}{c|}{$n=20$} & \multicolumn{3}{c|}{$n=50$} & \multicolumn{3}{c|}{$n=100$} & \multirow{2}*{\#Par.}\\
     & Obj. & Gap & Time & Obj. & Gap & Time & Obj. & Gap & Time  \\
    \midrule
     POMO, no aug.  & 6.19 & 1.09\% & 1s & 10.61 & 1.52\% & 3s & 16.02 & 1.08\% & 14s & 1.254M \\
     Ours (Full), no aug. & 6.19 & 1.08\% & 1s & 10.56 & 0.98\%& 3s & 16.02 & 1.06\% & 14s& 1.254M \\ 
     Ours (Inside Tuning), no aug. & 6.21 & 1.34\% & 1s & 10.64 & 1.80\%& 3s & 16.08 & 1.47\% & 16s& 0.256M \\   
     Ours (Side Tuning), no aug. & 6.25 & 1.94\% & 1s & 10.68 & 2.15\%& 3s & 16.27 & 2.66\% & 15s& 0.331M \\
    Ours (LoRA), no aug. & 6.24 & 1.80\% & 1s & 10.67 & 2.01\%& 3s & 16.19 & 2.15\% & 16s& \textbf{0.122M} \\
                  \cmidrule(lr){1-11}
     POMO, $\times8$ aug. & 6.14 & 0.29\% & 7s & 10.46 & 0.02\%& 22s & 15.86 & 0.06\% & 2m& 1.254M \\     
     Ours (Full), $\times8$ aug. & \textbf{6.13} & \textbf{0.00\%} & 7s & \textbf{10.46} & \textbf{0.00\%} & 22s & \textbf{15.85} & \textbf{0.00\%} & 2m& 1.254M \\ 
     Ours (Inside Tuning), $\times8$ aug. & 6.17 & 0.41\% & 7s & 10.51 & 0.53\%& 23s & 15.91 & 0.39\% & 2m& 0.256M \\ 
    
     Ours (Side Tuning), $\times8$ aug. & 6.15 & 0.44\% & 7s & 10.52 & 0.62\%& 23s & 16.06 & 1.33\% & 2m& 0.331M \\
    Ours (LoRA), $\times8$ aug. & 6.16 & 0.49\% & 7s & 10.53 & 0.67\%& 23s & 15.99 & 0.88\% & 2m& \textbf{0.122M}
     \\   
    
   
    \midrule
    \bottomrule
  \end{tabular}}
  \end{center}
  \vspace{-1mm}
  \caption{Inference results with POMO.}
  \vspace{2mm}
   \label{pomo1} 
\end{table*}


\begin{figure*}[!ht]
\vspace{-3mm} 
\centering
\begin{tabular}{@{}c@{}c@{}c@{}c@{}}
  {\includegraphics[width=0.33\linewidth]{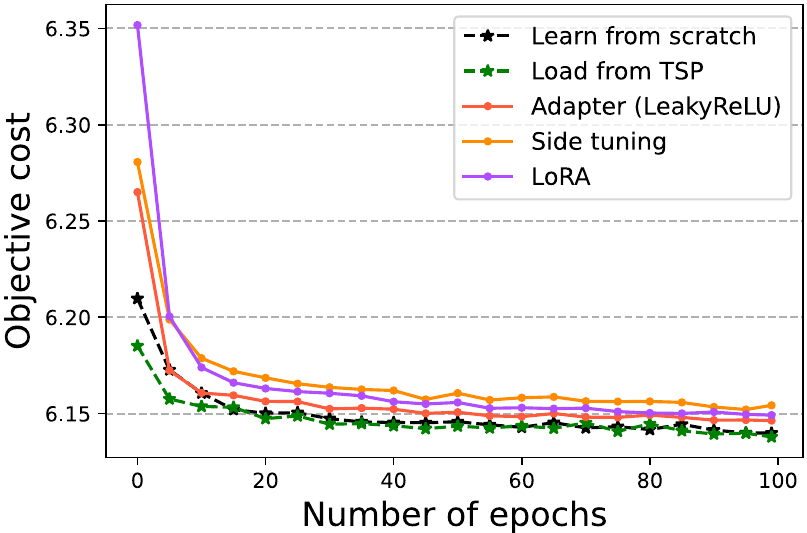}}&
  {\includegraphics[width=0.33\linewidth]{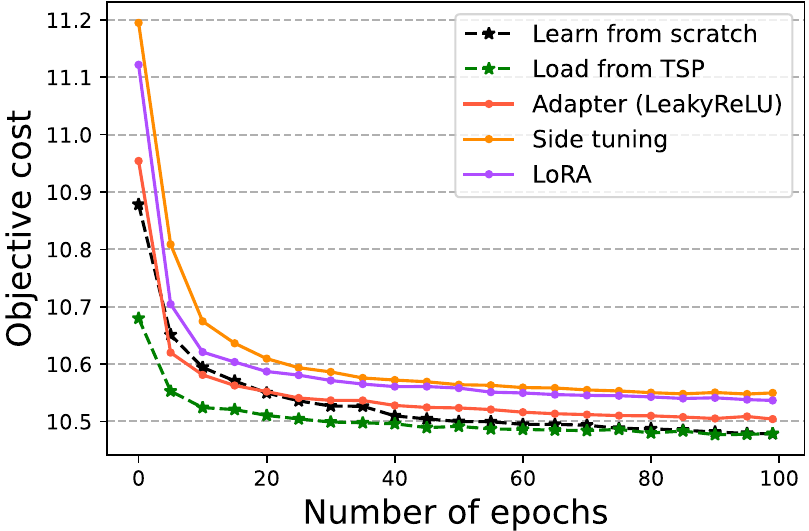}}&
  {\includegraphics[width=0.33\linewidth]{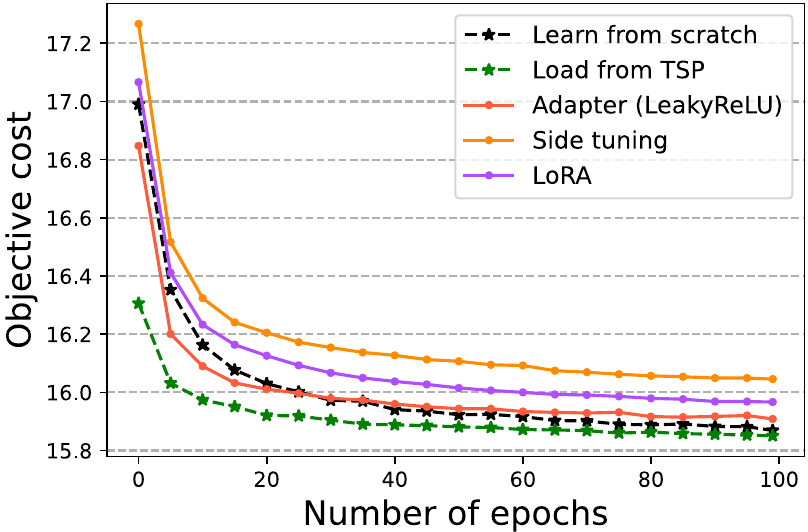}}&
\end{tabular}
\vspace{-2mm} 
\caption{Training curves with POMO on CVRP $(n=20, 50, 100)$, respectively.}
\label{fig:pomo} 
\vspace{-3mm}
\end{figure*}

\subsection{Versatility}
\label{sec:versa}
The cross-problem learning is versatile to different deep models for VRPs. We further deploy it with POMO~\cite{kwon2020pomo}, another Transformer based neural network which is developed from AM~\cite{kool2018attention} and achieves the state-of-the-art performance in solving TSP and CVRP. 
More details of POMO can be found in Appendix~C. 
Specifically, we keep all the experimental settings with the same hyperparameters for training the original POMO.
For our fine-tuning methods, we train POMO with the batch size 64, and the instances generated on the fly. For each epoch, we train 100,000 instances with them divided into batches. The deep model is trained with 100 epochs, when we observe that the training curves already converge. For the learning rate, we set $\eta=10^{-4}$ for the problem size $n=20$, and $\eta=10^{-5}$ for $n=50, 100$ to avoid overfitting in the training process. As shown in Figure~\ref{fig:pomo}, the full fine-tuning generally converges faster and better than POMO trained from scratch. Furthermore, the inside tuning attains close performance to POMO, saving 79.59\% parameters (i.e., 0.256M vs. 1.254M), which is more significant than the saving with AM (i.e., 71.33\%).


We demonstrate the results of POMO and our methods by two inference settings, i.e., with and without the instance augmentation (denoted as no aug. and $\times 8$ aug., respectively.)~\cite{kwon2020pomo}. We evaluate all methods on 10,000 CVRP instances for each of the sizes $n=20, 50, 100$. We list the average objective value, average gap (to the best solution), total inference time and the number of trained parameters in Table \ref{pomo1}.
As shown, our method with the full fine-tuning consistently performs better than POMO across all problem sizes. 
In particular, when the number of nodes $n$ is small, i.e., $n=20$, POMO finds solutions with an average gap of 0.29\% compared with the full fine-tuning method.
On the other hand, our method with the inside tuning performs slightly worse than POMO (e.g. 0.06\% vs. 0.39\% for $n=100$).  
This indicates that the adapter-based methods are able to achieve comparable performance with much fewer parameters for different deep learning backbones. 
As for the side tuning method, its performance is comparable to the inside tuning for $n=$20, 50, but becomes drastically inferior for $n$=100. 
Compared with side tuning, LoRA achieves comparable performance for $n=$20, 50 and better performance (i.e., 0.88\% vs. 1.33\%) for $n=100$ by saving 90.27\% (i.e., 0.122M vs 1.254M) parameters. To further evaluate our methods, we apply them to solve benchmark instances in CVRPLIB. The results show that the full fine-tuning significantly outperforms POMO on large-scale instances with sizes up to 1000. Detailed results are presented in Appendix~D.

\section{Conclusion and Future Work}
\label{conclusion}
This paper presents a cross-problem learning method for solving VRPs, which assists heuristics training for downstream VRPs by using the pre-trained Transformer for a basic VRP. We modularize Transformers for VRPs based on the backbone (Transformer) for TSP, and propose different fine-tuning methods to train the backbone or problem-specific modules in the target VRP. Results show that the full fine-tuning significantly outperforms Transformer trained from scratch, and the adapter-based fine-tuning delivers comparable performance, with far fewer parameters. We also verify the favorable cross-distribution application and versatility of our method, along with the efficacy of key components. 
In the future, we will improve our methods with more advanced techniques such as neural architecture search. We also plan to apply our method to solve other COPs, such as job shop scheduling problem or bin packing problem.

\newpage




\section*{Acknowledgments}
This research is supported by the National Research Foundation, Singapore under its AI Singapore Programme (AISG Award No: AISG3-RP-2022-031), and the Singapore Ministry of Education (MOE) Academic Research Fund (AcRF) Tier 1 grant.




\pdfpagewidth=8.5in
\pdfpageheight=11in




\definecolor{hhhh}{RGB}{51,107,158}
\definecolor{greyC}{RGB}{180,180,180}
\definecolor{greyL}{RGB}{235,235,235}
\definecolor{shadecolor}{rgb}{0.92,0.92,0.92}


\urlstyle{same}







\pdfinfo{
/TemplateVersion (IJCAI.2024.0)
}






\appendix
\section*{Appendix}
In this appendix, we first present the problem definitions of PCTSP and CVRP. 
To show the versatility of our cross-problem learning, we further investigate whether the cross-problem learning can effectively solve VRPs in the context of varying distributions.
Then we introduce another commonly used neural heuristic POMO, and provide additional evaluation results with POMO taken as the backbone Transformer for cross-problem learning. 
Finally, we test the generalization of the cross-problem learning on benchmark instances.

\section{PCTSP and CVRP}

\noindent\textbf{Prize Collecting TSP (PCTSP)}~\cite{balas1989prize}. The nodes in PCTSP are not only featured by coordinates $\{x_i\}_{i=0}^n$, but also prizes $\{p_i\}_{i=1}^n$ and penalties $\{\beta_i\}_{i=1}^n$. Specially, the depot node has no prize and penalty, i.e., $p_0=\beta_0=0$. We optimize the tour in PCTSP to minimize the total length plus the total penalty for the nodes that are not traversed. A feasible tour should start at the depot and finally end at depots, and guarantee at least a predefined total prize is collected in the tour. Besides, each node should be traversed at most once. 

\noindent\textbf{Capacitated Vehicle Routing Problem (CVRP)}~\cite{toth2014vehicle}. The nodes in CVRP are not only featured by coordinates $\{x_i\}_{i=0}^n$ but also demands $\{\delta_i\}_{i=1}^n$. Specially, $\delta_0=0$ for the depot node. The solution to CVRP is a tour that comprises multiple subtours. Each subtour starts at the depot, visits a subset of customers in sequence, and ends at the depot. A feasible tour should also guarantee that: 1) each customer is visited exactly once in the tour; 2) the total demand of customers on each subtour should not be larger than the predefined vehicle capacity.

\section{Cross-Distribution Application}
\begin{figure*}[htb!]
\centering
\begin{tabular}{@{}c@{}c@{}}
  \hspace{-3mm}
  {\includegraphics[width=0.45\linewidth]{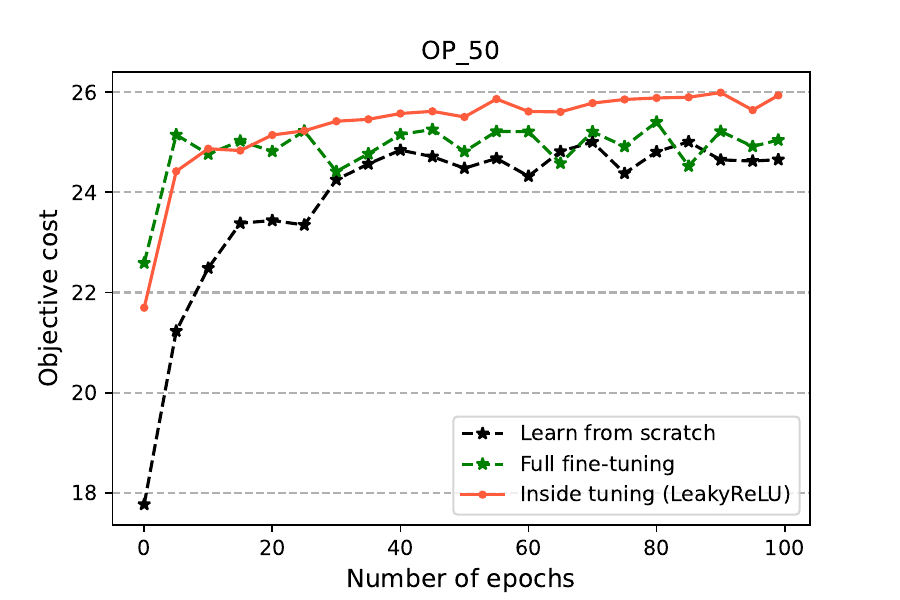}}&\hspace{10mm}
  {\includegraphics[width=0.45\linewidth]{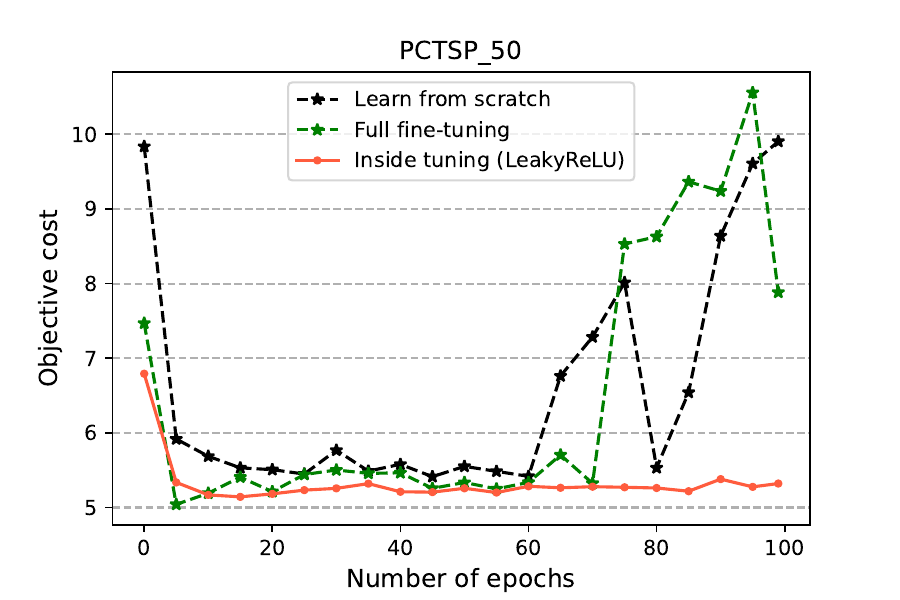}}
\end{tabular}
\vspace{-2mm} 
\caption{Cross-distribution application.}
\label{fig:distribution} 
\vspace{-0mm}
\end{figure*}

While the existing neural heuristics for VRP are often trained with Uniform instances, one question is if the pre-trained backbone for TSP would be generalizable to help the training for downstream VRPs with different distributions. To answer it, we apply the backbone pre-trained with Uniform TSP instances in the full fine-tuning and inside tuning for OP, PCTSP ($n$=50) instances from Gaussian(0, 1). Figure~\ref{fig:distribution} reveals that the backbone trained from Uniform distribution well assists in the training for another distribution. While the full fine-tuning and training from scratch are overfitted on PCTSP, the side tuning significantly outstrips them with faster and better convergence.  It implies our methods can save training overhead not only at a cross-problem level but also at a cross-distribution level. We would like to note that the proposed cross-problem learning method is different from current generalization techniques for neural heuristics of VRP, which rely on additional heavy optimization or search approaches for each problem or even each instance, e.g., distributionally robust optimization in ~\cite{jiang2022learning}, meta-learning in~\cite{qiu2022dimes}, active search in~\cite{hottung2022efficient}. In contrast, our method focuses more on how to deliver a better-performing Transformer for the target distribution during the original DRL training. It simply leverages the pre-trained model as a warm start, and saves massive manual work to specialize additional techniques for each VRP, which means a more sustainable training procedure. Moreover, it is straightforward to apply the current techniques (proposed for AM trained from scratch) on top of our method to potentially improve their performance. We will leave such attempts in our future work.

\begin{algorithm*}[ht]
  \small
  \caption{Training Algorithm for POMO}
  \label{alg:pomo}
  
  {\bf Input:} Instances for training $\{\mathcal{G}_k\}_{k=1}^K$, number of starting nodes $n$, steps $T$, batch size $B$; \\
  {\bf Output:} Trained model $p_\theta^{\star}$;
\begin{algorithmic}[1]
\WHILE{$t<T$}
    \STATE $\mathcal{G}_j \gets$ sample instance($\{\mathcal{G}_k\}_{k=1}^K$) \quad $\forall j \in \{1, \ldots, B\}$
    
    \STATE $\{v^j_1, v^j_2, \cdots, v^j_n\} \gets$ selectstartingnodes($\mathcal{G}_j$) \quad $\forall j \in \{1, \ldots, B\}$
    
    \STATE $\pi^m_j \gets p_\theta(v^j_m, \mathcal{G}_j, p_\theta) \quad \forall j \in \{1, \ldots, B\},\,\, \forall m \in \{1, \ldots, n\}$
    
    \STATE $b(\mathcal{G}_j) \gets \frac{1}{n} \sum_{m=1}^n \mu(\pi^m_j)$ \quad $\forall m \in \{1, \ldots, M\}$ 
    
    \STATE $\nabla_{\theta}\mathcal{L}(\theta) \gets \frac{1}{Bn} \sum_{m=1}^n \sum_{j=1}^B (\mu(\pi^m_j)-b(\mathcal{G}_j)) \nabla_{\theta}\log p_{\theta}(\pi^m_j|\mathcal{G}_j)$
\ENDWHILE
\end{algorithmic}
\end{algorithm*}

\begin{table*}[!ht] \small
\begin{center}
\caption{Generalization to benchmark instances (without augmentation).}
\begin{tabular}{cccccccccccc}
\toprule
\midrule
\multicolumn{2}{c|}{\multirow{2}{*}{}}             & \multicolumn{2}{c}{\multirow{2}{*}{\begin{tabular}[c]{@{}c@{}}POMO \\ no aug.\end{tabular}}} & \multicolumn{2}{c}{\multirow{2}{*}{\begin{tabular}[c]{@{}c@{}}Full fine-tuning\\ no aug.\end{tabular}}} & \multicolumn{2}{c}{\multirow{2}{*}{\begin{tabular}[c]{@{}c@{}} Inside Tuning\\ no aug.\end{tabular}}} & \multicolumn{2}{c}{\multirow{2}{*}{\begin{tabular}[c]{@{}c@{}}Side Tuning\\ no aug.\end{tabular}}} & \multicolumn{2}{c}{\multirow{2}{*}{\begin{tabular}[c]{@{}c@{}}LoRA\\ no aug.\end{tabular}}} \\
\multicolumn{2}{c|}{}                              & \multicolumn{2}{c}{}                                                                          & \multicolumn{2}{c}{}                                                                                & \multicolumn{2}{c}{}                                                                                         & \multicolumn{2}{c}{}     & \multicolumn{2}{c}{}                                                                                   \\ \hline
Instance             & \multicolumn{1}{c|}{Opt.}   & Obj.                                         & Gap                                            & Obj.                                             & Gap                                              & Obj.                                                 & Gap       & Obj.                                                 & Gap                                              & Obj.                                        & \multicolumn{1}{c}{Gap}                                      \\
X-n101-k25           & \multicolumn{1}{c|}{27591}  & \textbf{29752} & \textbf{7.83\%}  & 29948           & 8.54\%           & 30236              & 9.59\%              & 31987     & 15.93\%     & 29811    & 8.04\%               \\
X-n153-k22           & \multicolumn{1}{c|}{21220}  & \textbf{24635} & \textbf{16.09\%} & 24816           & 16.95\%          & 24924              & 17.46\%             & 27859     & 31.29\%      & 25278    & 19.12\%               \\
X-n200-k36           & \multicolumn{1}{c|}{58578}  & 62030          & 5.89\%           & \textbf{61759}  & \textbf{5.43\%}  & 62245              & 6.26\%              & 62322     & 6.39\%       &  64618   & 10.31\%               \\
X-n251-k28           & \multicolumn{1}{c|}{38684}  & \textbf{40682} & \textbf{5.16\%}  & 40809           & 5.49\%           & 41843              & 8.17\%              & 43037     & 11.25\%           &  42665   &  10.29\%         \\
X-n303-k21           & \multicolumn{1}{c|}{21736}  & 23399          & 7.65\%           & \textbf{23147}  & \textbf{6.49\%}  & 24798              & 14.09\%             & 26066     & 19.92\%         &   26635  & 22.54\%           \\
X-n351-k40           & \multicolumn{1}{c|}{25896}  & 29096          & 12.36\%          & \textbf{28529}  & \textbf{10.17\%} & 30424              & 17.49\%             & 37171     & 43.54\%         &  33021   &  13.49\%          \\
X-n401-k29           & \multicolumn{1}{c|}{66154}  & \textbf{71663} & \textbf{8.33\%}  & 71954           & 8.77\%           & 80054              & 21.01\%             & 72970     & 10.30\%          &  77854   & 17.69\%         \\
X-n459-k26           & \multicolumn{1}{c|}{24139}  & 28570          & 18.36\%          & \textbf{27211}  & \textbf{12.73\%} & 30135              & 24.84\%             & 33171     & 37.42\%         &  33387   &  38.31\%          \\
X-n502-k39           & \multicolumn{1}{c|}{69226}  & 76834 & 10.99\% & 78660           & 13.63\%          & 91598              & 32.32\%             & 197162    & 184.81\%        & \textbf{75973}    &  \textbf{9.75\%}          \\
X-n548-k50           & \multicolumn{1}{c|}{86700}  & 94538          & 9.04\%           & \textbf{92805}  & \textbf{7.04\%}  & 100005             & 15.35\%             & 178352    & 105.71\%        & 93480    &  7.82\%          \\
X-n599-k92           & \multicolumn{1}{c|}{108451} & 121919         & 12.42\%          & \textbf{119353} & \textbf{10.05\%} & 127411             & 17.48\%             & 134381    & 23.91\%        & 121525    &  12.06\%           \\
X-n655-k131          & \multicolumn{1}{c|}{106780} & 168843         & 58.12\%          & \textbf{116113} & \textbf{8.74\%}  & 125496             & 17.53\%             & 327427    & 206.64\%        & 121619    &  13.90\%          \\
X-n701-k44           & \multicolumn{1}{c|}{81923}  & 90180          & 10.08\%          & \textbf{89118}  & \textbf{8.78\%}  & 99045              & 20.90\%             & 102428    & 25.03\%        &  91171   &   11.29\%          \\
X-n749-k98           & \multicolumn{1}{c|}{77269}  & 88728          & 14.83\%          & \textbf{88031}  & \textbf{13.93\%} & 92045              & 19.12\%             & 110432    & 42.92\%       &  114705   &  48.45\%           \\
X-n801-k40           & \multicolumn{1}{c|}{73311}  & 83730          & 14.21\%          & \textbf{80937}  & \textbf{10.40\%} & 104672             & 42.78\%             & 231414    & 215.66\%      &  83244   &  13.55\%            \\
X-n856-k95           & \multicolumn{1}{c|}{88965}  & \textbf{99084} & \textbf{11.37\%} & 100005          & 12.41\%          & 106479             & 19.69\%             & 264540    & 197.35\%      & 106628     &  19.85\%            \\
X-n895-k37           & \multicolumn{1}{c|}{53860}  & 64046          & 18.91\%          & \textbf{62681}  & \textbf{16.38\%} & 66558              & 23.58\%             & 116219    & 115.78\%       & 68092    &  26.42\%           \\
X-n957-k87           & \multicolumn{1}{c|}{85465}  & 104280         & 22.01\%          & \textbf{98192}  & \textbf{14.89\%} & 100671             & 17.79\%             & 378077    & 342.38\%      & 104245    &  21.97\%            \\
X-n1001-k43          & \multicolumn{1}{c|}{72355}  & 85406          & 18.04\%          & \textbf{82381}  & \textbf{13.86\%} & 97784              & 35.14\%             & 147965    & 104.50\%     & 83244     &   15.05\%            \\ 
\midrule
\bottomrule
\end{tabular}
\end{center}
\vspace{-2mm} 
\end{table*}

\begin{table*}[!ht] 
\vspace{1.5mm} 
\begin{center}
  \renewcommand\arraystretch{1.05}
\caption{Generalization to benchmark instances (with augmentation).}
\label{pomo2}
  \resizebox{\textwidth}{!}{
\begin{tabular}{ll|cccccccccc}
\toprule
\midrule
\multicolumn{2}{c|}{\multirow{2}{*}{}}             & \multicolumn{2}{c}{\multirow{2}{*}{\begin{tabular}[c]{@{}c@{}}POMO \\ $\times8$ aug.\end{tabular}}} & \multicolumn{2}{c}{\multirow{2}{*}{\begin{tabular}[c]{@{}c@{}}Full fine-tuning\\ $\times8$ aug.\end{tabular}}} & \multicolumn{2}{c}{\multirow{2}{*}{\begin{tabular}[c]{@{}c@{}}Inside Tuning\\ $\times8$ aug.\end{tabular}}} & \multicolumn{2}{c}{\multirow{2}{*}{\begin{tabular}[c]{@{}c@{}}Side Tuning\\ $\times8$ aug.\end{tabular}}} & \multicolumn{2}{c}{\multirow{2}{*}{\begin{tabular}[c]{@{}c@{}}LoRA\\ $\times8$ aug.\end{tabular}}}\\
\multicolumn{2}{c|}{}                              & \multicolumn{2}{c}{}                                                                          & \multicolumn{2}{c}{}                                                                                & \multicolumn{2}{c}{}                                                                                         & \multicolumn{2}{c}{}                                                                                       \\ \hline
Instance             & \multicolumn{1}{c|}{Opt.}   & Obj.                                         & Gap                                            & Obj.                                             & Gap                                              & Obj.                                                 & Gap                                                   & Obj.                                        & Gap                 & Obj.                                        & Gap                     \\

X-n101-k25   & 27591  & \textbf{29337}    &  \textbf{6.33\%}    & 29439                &   6.70\%        & 29544                &   7.08\%                 & 30739               &   11.41\%       &29525   &     7.01\%    \\
X-n153-k22   & 21220  & 23501             &   10.75\%    & \textbf{23291}       & \textbf{9.76\%}          & 23420                &10.37\%                    & 24415               &  15.06\%     &24125   &       13.69\%     \\
X-n200-k36   & 58578  & 61672             &  5.28\%     & \textbf{61668}       &  \textbf{5.28\%}         & 62128                &  6.06\%                  & 62322               &  6.39\%          &62776   &    7.17\%   \\
X-n251-k28   & 38684  & \textbf{40682}    &  \textbf{5.16\%}     & 40809                &   5.49\%        & 41286                &  6.73\%                  & 41564               &   7.44\%        & 41457  &   7.17\%     \\
X-n303-k21   & 21736  & 23306             &   7.22\%    & \textbf{23104}       &    \textbf{6.29\%}       & 23477                &    8.01\%                & 24931               &     14.70\%    & 25212  &    15.99\%      \\
X-n351-k40   & 25896  & \textbf{28382}    &  \textbf{9.60\%}     & 28500                &  10.06\%         & 30424                &   17.49\%                 & 32016               &    23.63\%    & 31041  &   19.87\%        \\
X-n401-k29   & 66154  & 70685             &   6.85\%    & \textbf{70142}       &  \textbf{6.03\%}         & 70577                &    6.69\%                & 72970               &   10.30\%        & 73385  &  10.93\%      \\
X-n459-k26   & 24139  & 27800             &   15.17\%    & \textbf{27211}       &   \textbf{12.73\%}        & 29506                &     22.23\%               & 33171               &   37.42\%     & 30653  &  26.99\%         \\
X-n502-k39   & 69226  & 76623             & 10.69\%      & \textbf{73912}       &    \textbf{6.77\%}       & 77955                &  12.61\%                  & 153176              &  121.27\%     & 75664  &  9.30\%          \\
X-n548-k50   & 86700  & 92683             &   6.90\%    & \textbf{92565}       &  \textbf{6.76\%}         & 95527                &      10.18\%              & 178352              &   105.71\%    & 93480  &   7.82\%         \\
X-n599-k92   & 108451 & \textbf{118166}   &  \textbf{8.96\%}     & 119353               & 10.05\%          & 122701               &   13.14\%                 & 134381              & 23.91\%       & 121525  &   12.06\%        \\
X-n655-k131  & 106780 & 156973            &    47.01\%   & \textbf{113548}      &   \textbf{6.34\%}        & 117447               &  9.99\%                  & 277623              &    160.00\%     & 120505  &  12.85\%        \\
X-n701-k44   & 81923  & 89076             &    8.73\%   & \textbf{88959}       &   \textbf{8.59\%}        & 91669                &   11.90\%                 & 102428              &    25.03\%        & 91023  & 11.11\%      \\
X-n749-k98   & 77269  & \textbf{85901}    &   \textbf{11.17\%}    & 86059                &      11.38\%     & 90747                & 17.44\%                   & 103398              &   33.82\%      & 95376  &   23.43\%       \\
X-n801-k40   & 73311  & 82336             &   12.31\%    & \textbf{80806}       &   \textbf{10.22\%}        & 84046                &  14.64\%                  & 219087              &     198.85\%     & 83244  &  13.55\%       \\
X-n856-k95   & 88965  & \textbf{98874}    & \textbf{11.14\%}      & 99255                &  11.57\%         & 101653               &  14.26\%                  & 264540              &     197.35\%     &  106161 & 19.33\%        \\
X-n895-k37   & 53860  & 62544             &     16.12\%  & \textbf{61771}       &  \textbf{14.69\%}         & 65553                &  21.71\%                  & 91850               &70.53\%            & 64780  &  20.27\%     \\
X-n957-k87   & 85465  & 98572             & 15.34\%     & \textbf{95891}       &   \textbf{12.20\%}        & 99249                &     16.13\%               & 238819              & 179.43 \%         & 104245  & 21.97\%        \\
X-n1001-k43  & 72355  & 83823             &   15.85\%    & \textbf{81888}       &   \textbf{13.18\%}       & 86020                &  18.89\%                  & 115267              &    59.30\%       & 84915  &  17.36\%      \\ 

\midrule
\bottomrule
\end{tabular}}
\end{center}
\vspace{1.5mm} 
\end{table*}

\section{POMO}
\label{app:pomo}
We introduce the main extensions of POMO~\cite{kwon2020pomo} on top of AM~\cite{kool2018attention}. POMO mainly enhances AM in two aspects. On the one hand, POMO leverages the multiple starting nodes for each TSP or CVRP instance to explicitly enforce the policy network to search multiple optima. The training objective in the DRL algorithm (i.e. REINFORCE) is also changed to benefit the policy learning. On the other hand, POMO resorts to the data augmentation to further increase the exploration of better solutions for each instance. We detail these two aspects as follows.

\noindent\textbf{Multiple Starting Nodes.} Let's assume a solution to a TSP instance with 5 nodes is $(v_1, v_2, v_3, v_4, v_5)$. The policy should also learn that  $(v_2, v_3, v_4, v_5, v_1)$, $(v_3, v_4, v_5, v_1, v_2)$, $\ldots$ represent the same solution. According to this point, POMO explicitly provides multiple starting nodes for each instance during the policy learning, so that the learned policy is able to solve the same instance from different angles. To well consider the multiple starting nodes, POMO changes the objective function in REINFORCE (see Section 3.2 in the main paper) as below:
\begin{equation*}
    \label{eq:pomo_reinforce}
    \nabla_{\theta} \mathcal{L}(\theta|\mathcal{G}) = \mathbb{E}_{\substack{p_{\theta}(\pi^m|\mathcal{G}),\\ m\in (1,\cdots,n)}} [(\mu(\pi^m)-b(\mathcal{G})) \nabla_{\theta}\log p_{\theta}(\pi^m|\mathcal{G})],
\end{equation*}
where $\pi^m$ is the solution to the instance $\mathcal{G}$ with the $m$-th node in $\{v_i\}_{i=1}^n$ as the starting node, and $n$ denotes the number of different starting nodes which is essentially equal to the number of customer nodes in TSP and CVRP. For the baseline function, POMO defines $b(\mathcal{G})=\frac{1}{n}\sum_{m=1}^n \mu(\pi^m)$, which intuitively encourages the better solutions for each instance in comparison to the average performance of solutions derived by different starting nodes. In this way, the learned policy can solve the same instance from different angles (i.e. starting nodes) with all the derived solutions gaining favorable performance, as they are improved by self-playing with their (better and better) average performance. 

\noindent\textbf{Instance Augmentation.} The data augmentation for TSP or CVRP instances is executed by considering the symmetry of the instances and their solutions. Given the coordinates of nodes in an instance $\{x_i\}_{i=0}^n$ in the unit square $[0,1]\times[0,1]$, it could be extended as different instances that share the same solution set. For example, $\{x_i^0, x_i^1\}_{i=0}^n$ could be transformed into $\{x_i^1, x_i^0\}_{i=0}^n$; $\{x_i^0, 1-x_i^1\}_{i=0}^n$ could be transformed into $\{x_i^1, 1-x_i^0\}_{i=0}^n$. POMO uses these extra instances in the inference for each instance, which again provides the potentially different solutions to the same instance, so as to enhance the exploration for better solutions.

The pseudocode of the training algorithm for POMO is presented in Algorithm \ref{alg:pomo}. We refer the interested readers for more details in the original paper~\cite{kwon2020pomo}.

\section{Generalization to Benchmark Instances}

To further evaluate our models trained by the cross-problem learning, we apply them to directly solve benchmark instances from the dataset CVRPLIB.\footnote{\url{http://vrp.atd-lab.inf.puc-rio.br/index.php/en/}} Specifically, we choose random instances with sizes $n \in [101, 1001]$ from the instance group proposed in~\cite{uchoa2017new}. The original and our POMO models trained for CVRP of size $n=100$ are used to solve the instances, with the instance augmentation during inference. We report the objective value and the gap to the optimal solution for each instance in Table~\ref{pomo2}. As shown, our method with the full fine-tuning outperforms POMO for most instances, where the best results for each instance are marked in \textbf{bold}. Moreover, we observe that our method with the inside tuning is comparable to POMO, and even outperforms POMO on X-n153-k22, X-n401-k29 and X-n655-k131. In summary, the benchmark results verify that the full fine-tuning and inside tuning deliver reliable models, which generalize fairly well to instances that are totally different from the training instances.

\bibliographystyle{named}
\bibliography{ijcai24}

\end{document}